\documentclass[conference]{IEEEtran}
\IEEEoverridecommandlockouts
\usepackage{cite}
\usepackage{amsmath,amssymb,amsfonts}
\usepackage{algorithmic}
\usepackage{graphicx}
\usepackage{textcomp}
\usepackage{xcolor}
\def\BibTeX{{\rm B\kern-.05em{\sc i\kern-.025em b}\kern-.08em
    T\kern-.1667em\lower.7ex\hbox{E}\kern-.125emX}}
\begin{document}

\title{The Right Tool for the Job: Open-Source Auditing Tools in Machine Learning\\
}

\author{\IEEEauthorblockN{Cherie M Poland}
\IEEEauthorblockA{\textit{Complex Adaptive Systems Research} \\
Spring, Texas, USA \\
cmpoland1030@gmail.com \\
cheriem@vt.edu\\
ORCID:0000-0002-6345-649X}

}

\maketitle

\begin{abstract}
In recent years, discussions about fairness in machine learning, AI ethics and algorithm audits have increased. Many entities have developed framework guidance to establish a baseline rubric for fairness and accountability. However, in spite of increased discussions and multiple frameworks, algorithm and data auditing still remain difficult to execute in practice.
Many open-source auditing tools are available, but users aren't always aware of the tools, what they are useful for, or how to access them. Model auditing and evaluation are not frequently emphasized skills in machine learning.
There are also legal reasons for the proactive adoption of these tools that extend beyond the desire for greater fairness in machine learning. There are positive social issues of public perception and goodwill that matter in our highly connected global society. 
Greater awareness of these tools and the reasons for actively utilizing them may be helpful to the entire continuum of programmers, data scientists, engineers, researchers, users and consumers of AI and machine learning products. It is important for everyone to better understand the input and output differentials, how they are occurring, and what can be done to promote FATE (fairness, accountability, transparency, and ethics) in machine- and deep learning.
The ability to freely access open-source auditing tools removes barriers to fairness assessment at the most basic levels of machine learning.
This paper aims to reinforce the urgent need to actually use these tools and provides motivations for doing so. The exemplary tools highlighted herein are open-source with software or code-base repositories available that can be used immediately by anyone worldwide.

\end{abstract}

\begin{IEEEkeywords}
algorithm analysis, problem complexity, auditing, fairness, accountability, transparency, ethics, tools, open-source, machine learning, law.
\end{IEEEkeywords}

\section{Introduction}
As a society, we are tossed around on a sea of competing ideas, backed by statistics, data, and algorithms. The decisions made by algorithms rely on the data used as input and the manner in which the algorithms are configured and programmed by humans. Ultimately, however, all decisions are human decisions, regardless of the manner or means by which the data are filtered, manipulated and wielded. Humans select the filters, manipulate the data, and wield the algorithms. That means that humans can also ultimately be held responsible for the outcome of those decisions, whether directly made by a human brain or by extension, an algorithmic matrix.
It is widely known that algorithms can encompass, express, and unveil intentional or unconscious bias, including bias intrinsically present in the data, and manipulation by competing interests or those with agendas.
The critical need for fairness, accountability, transparency, and ethics (FATE) in machine learning has increased dramatically in recent years as the output of systems has shown the disparities in data and algorithms in every field of endeavor. This increase in awareness has also triggered legal concerns that have evolved into rubrics for ethical use of algorithmic tools, but real-world legal concerns are being raised and governmental regulators around the world are being educated on remedies that may be needed if the industry does not better police itself.

Audits are evaluations with expectations of accountability \cite{b1}. Algorithmic auditing is not a one-size-fits all solution. Different methods of auditing are required in order to search for and test both the data and the algorithms for underlying or hidden bias and attributes. Methods encompassing anomaly detection and pattern recognition are required. No single auditing method will likely uncover all potential issues. The breadth and scope of auditing methods necessitates creating an algorithmic auditing toolkit that one may use to deeply examine data and algorithms with various degrees of granularity.

\section{Why Auditing is Important}
Fairness is an important concern in terms of trust and reliability of the decision-making process involving machine learning models. Each model has specific requirements for the type of data it will operate over. These necessitate different pre-processing techniques. However, in order to provide a consistent comparison across models, each model must receive similar input. Data are manipulated by pre-processing it to generate similar types of inputs that models can handle. However, this data manipulation can also create bias in datasets as it is a deliberate method of focusing the analytical lens on a particular set of data. Data may also be biased because it is a collection of human-generated information.
There is an evidence-based assumption that bias is bad. There is a societal need to be accountable, transparent, and fair, especially because the outputs of algorithms are not always able to be understood or interpreted by humans. Algorithmic output and decisions are also susceptible to inaccuracies, may be adversarial to human rights or other fundamental rights, and may present legal challenges such as data privacy and unauthorized use or sale of data by third-parties. Auditing algorithms may not resolve all issues of concern, but audits are part of the solution in providing more ethical accountability for very powerful machine learning software.

\subsection{Model Standards}
The implementation of reproducible and standardized techniques has led industry to establish guidelines for fairness, accountability, and transparency in machine learning models \cite{b2}. However, the implementation of these guidelines requires tools. The main objective of these tools is to provide a mechanism by which bias and other forms of concerning algorithmic output can be detected, measured, and reduced.

Various auditing tools and open-source libraries have been developed to provide a high level of functionality in machine learning and deep learning models.

Different models of analysis are used to extract different features from data. These features, their patterns, and anomalies are then analyzed to determine whether there are irregularities in the underlying data or in the algorithm used that need to be explored further.
The expectation of finding bias of some type in any given generic dataset should be neutral. While none is overtly expected, history shows that humans have both overt and unconscious bias and that bias may be reflected in unexpected places.

Pervasive use of high performing, 'black-box' models in sensitive areas like healthcare\cite{b3} and criminal justice systems raises ethical concerns and a critical need to audit these models for fairness. Recent works\cite{b4} and \cite{b5} have revealed disturbing bias in some machine learning models against minority groups. 

One way to audit these algorithms is to look inside the inner-workings of the algorithm, model, and datasets to understand how the underlying algorithms, data, data pre-processing, input features, feature engineering, tuning, and optimization are combined to arrive at the model output. However, due to proprietary reasons, the inner workings of machine learning models may not be publicly available. One possible solution to this dilemma is that post-hoc interpretability techniques can serve as useful auditing tools.

\subsection{The Law}
There are also growing legal reasons to be concerned about fairness in machine learning. Legal standards differ in different countries and what may be considered fair on one country's legal system may not be fair in another. Much of the law related to computer science is related to privacy and access to data and systems. In the United States employment laws related to disparate impact protect the public from the harm posed by biased algorithms, in some cases, but only once that disparate impact is discovered and challenged. There are, however, important legal distinctions. Disparate impact, by itself, is not unlawful. Disparate treatment may be unlawful in certain situations, particularly when individuals with a legally recognized protected status are harmed. However, even in this circumstance, the burden of proof is very high, is is very fact-specific, and will be limited to the specific context in which it arises.
There are anti-discrimination laws that apply to access to public services, housing, lending, and the like, but US law is a negative limitation on government, not the private sector.

In other areas of the world, the laws are very different. The laws of the European Union tend to be more stringent than laws in the United States. The General Data Protection Regulation (GDPR) protects privacy rights as well as rights related to disparate treatment and discrimination. In Asia, some laws related to machine learning systems are more lax than those of the United States. The laws in many South America and African countries are limited insofar as machine learning is concerned. Many countries have data protection laws, but their degree of protections vary.

The lack of regulation at the present time is an opportunity for the field to place a greater emphasis on self-policing.  The more upset the global public becomes with algorithms that show unfairness, bias, disparate impact, disparate treatment, and differential classification that discriminates against one group over another, the more attention will be drawn to the field and the calls for governmental regulation will increase.

These laws and regulations may be sector-specific, such as requiring computer vision algorithms to comply with safety standards of other vehicles driven on public roadways. Or they may be broader in scope, such as algorithms related to credit scoring and the manner in which they affect lending interest rates as a surrogate of access to or denial of financial or real estate services.

Fairness in machine learning is generally related to fairness of decision outcomes (distributive fairness), under the law. This is why the area is often examined under the disparate impact and disparate treatment legal rubrics. However, fairness in machine learning is also related to procedural fairness, especially in settings where there may be a deprivation of a legally-guaranteed right or privilege, such as a criminal, employment, or financially-based setting. \cite{b6} This also applies to computer vision algorithms related to biometrics, many of which simply apply Euclidean distance metrics and classifiers. The manner in which facial recognition software operates is important because a 1:1 identification has far less risk of discrimination or misidentification that a 1:N or 1:many analysis where an unknown face is compared to a large database of known faces to determine the unknown person's identity.

In a private setting, legal concerns about fairness in machine learning are tied directly to concerns about risk-management and public perception. A corporation can be sued or can easily lose brand status because of using machine learning algorithms that are publicly considered to be unfair, even if that unfairness does not rise to the level of unlawfulness in a civil context or illegality in a criminal context.

Educating the public by publishing algorithmic auditing assessments and proof of the use of auditing tools will help mitigate some of the concerns about fairness in machine learning algorithms. Taking responsibility for the outcome of the algorithm and its use will promote public good will.

A worst-case scenario is for industry to do nothing until a legislative body decides to regulate machine learning technologies in a strict manner similar to other product liability laws. This is what is slowly occurring in the autonomous driver space. When vehicle manufacturers cannot control damage to property and lives due to faults in their algorithms, the entities supplying the software, the executives, and engineers responsible may be held legally accountable.

\section{Exemplary Areas of Concern}
At the 2021 Neural Information Processing Systems (NeurIPS) conference, the organizing committee instituted an Ethics Review process for papers submitted to the conference.
According to the conference Retrospective publication, 9122 total papers were submitted and reviewed and 265 were flagged for additional ethics review in one or more areas of concern.

Areas of concern included: Discrimination/Bias/Fairness Concerns, Inadequate Data and Algorithm Evaluation, Inappropriate Potential Applications and Impact (e.g. human rights concerns), Legal Compliance (e.g. GDPR, copyright, terms of use), Privacy and Security (e.g. consent), Responsible Research Practice (e.g. IRB, documentation, research ethics), and research integrity issues (e.g. plagiarism) \cite{b7}.

The ACL, CVPR, ICLR, and EMNLP conferences have also taken steps to establish ethical expectations for the research community including introducing ethics review processes \cite{b7}. Including fairness and ethics assessments in the underlying work and research methodologies and reporting the same in the papers, reviews, and conference presentations will boost confidence from industry, academia, government, business, the public, and end-users who have concerns about AI and machine learning fairness and safety.

In January 2022, the United States Internal Revenue Service (IRS) announced that it would soon require facial recognition technology in order to access its systems \cite{b8} and \cite{b9}. The IRS contracted with a third-party company, ID.me, to provide the service. The company also provides services to the US Department of Veterans Affairs and the US Social Security Administration.  However, the agency reversed its decision after large public outcries over the practice \cite{b10}.

Even more chilling data privacy concerns were uncovered following the public outcry that showed the IRS had consented to allowing ID.me to “disclose or share Non-Personally Identifiable Information with Third-Party Service Providers and Affiliates” \cite{b11}. This implies that the company was or intended to be a data broker and many data privacy advocates have become rightly concerned.

Initially, ID.me initially stated that it was only going to be using 1:1 facial recognition, but later clarified that the company also uses 1:N (1:many) technology, prompting privacy advocates to raise public concerns. \cite{b12}

The CEO founder of ID.me stated that the 1:many is only used during initial enrollment and that afterwards, 1:1 is used \cite{b13}. However, a New Jersey man was indicted by the US Department of Justice in January 2022 for deliberately circumventing the security measures of ID.me in order to further a scheme to defraud the California Employment Development Department (EDD), filing at least 78 fraudulent unemployment insurance claims with EED seeking Pandemic Unemployment Assistance and other benefits under the federal CARES Act. EDD has partnered with ID.me since at least October 2021 \cite{b13}.

How much better off would ID.me be if they could show that they or a neutral third-party entity has conducted an algorithm audit of their systems and processes and be able to provide proof of the same?
In situations where the algorithmic model and data are proprietary, the use of something like zero-knowledge proofs may be a way to affirmatively show the results of an algorithmic and data audit without actually opening the proprietary system for analysis to the public or in a public forum.

As issues of fairness in machine learning and AI ethics continue to expand, it becomes imperative for the community to be aware of the potential risk and legal implication and the need to be familiar with open-source and free auditing tools that can be used to insure integrity of data and algorithms used.

\section{Techniques}

In recent years several machine learning interpretability techniques have emerged that can be used to explain the machine learning model results in terms of individual input feature contributions. Current popular interpretability techniques can be categorized into post-hoc interpretability and intrinsic interpretability \cite{b14}. Post-hoc methods provide explanations of the model output after the model is trained. Some of the current state-of-art methods include feature attributions (LIME, SHAP), model and data analysis tools (What-If Tool, Fairness Indicators), gradient and concept based testing (TCAV, GRAD-CAM) and data point inspection methods (partial dependence plots, ablation testing). Intrinsic interpretability is achieved by using inherently interpretable machine learning models.

In many instances, deep learning models have been shown provide a higher degree of accuracy than the traditional machine learning models when the input consists of large, unstructured data. Similarly, sophisticated ensemble techniques like bagging and boosting are preferred to simpler, more interpretable decision trees as they achieve better performance with structured tabular data. However, there exists a trade-off between accuracy and interpretability. The complex architecture of these models make it hard for the end-user to interpret how the input features have contributed to the model output prediction.

Post-hoc interpretability techniques serve as viable model auditing tools as they enable us to understand individual feature contributions to the model output.

In certain scenarios the explanations provided by the post-hoc methods are unreliable. Using inherently interpretable models can overcome this drawback.

\section{Tools}
Data and algorithms come in a variety of flavors. This is also true of auditing tools. It is important to know and understand which tool is appropriate to use at each phase of the machine learning continuum.
The main objective with these tools is to detect, measure, and reduce unwanted bias.
Some of these tools also incorporate simulations and data visualization models as graphical representations that can be used to discern differences in datasets in order to help identify patterns or anomalies.
The open-source framework facilitates a direct comparison of algorithms.

A sampling of a wide variety of tools and toolkits with links to papers, models, tutorials, instructions for use, and code repositories are as follows.

\subsection{AI Fairness360}

The AI Fairness 360 toolkit (AIF360) is a comprehensive Python open-source package developed at IBM that contains over 70 fairness metrics and 10 bias mitigation algorithms. \cite{b15} It is designed to translate algorithmic research from the lab into the actual practice of domains as wide-ranging as finance, human capital management, healthcare, and education. \cite{b16} In some cases, AI360 toolkit requires datasets to be transformed into Binary-Label-Dataset. The protected attributes can the be selected for observational analysis. These attributes partition a data population into groups that have a difference in terms of parity received.

Prior to applying fairness metrics to the selected attributes, standard statistical analysis can also be applied to determine the target attribute.

Code and instructions can be found at \cite{b17}, \cite{b18} and \cite{b19}.

\subsection{LIME}

LIME (Local Interpretable Model-Agnostic Explanations) \cite{b20} is a popular perturbation-based interpretability technique that provides local explanations to black-box models. LIME trains a linear classifier that locally approximates the behavior of the black-box machine learning model. A perturbed dataset is created by feeding in variations of a single data point to the ML model and noting the corresponding outputs. LIME then trains an interpretable model (for example, a decision tree) on this new dataset of permuted samples, where the samples are weighted by their proximity to the sample of interest. The local interpretable model is then used to explain the predictions made by the machine learning model to be audited \cite{b21}.

The LIME results and explanations can serve as a helpful model auditing tool when the inner workings of the machine learning model are not revealed. The user can feed in data samples permuted around a sample of interest to the black-box model and store the outputs as a new dataset to train a more interpretable model. 
However, LIME suffers from a drawback of unclear coverage. The explanations provided by LIME are local in nature and the boundary to which these explanations apply is not well defined.

Code and instructions can be found at \cite{b22}.

\subsection{ANCHORS}

ANCHORS (High-Precision Model-Agnostic Explanations) is a model-agnostic system proposed by the authors of LIME that generates a set of 'if-then' rules as explanations for model output predictions and overcomes the drawbacks of LIME by providing a clear coverage of the data instances to which the explanations apply \cite{b23}.

Code and instructions can be found at \cite{b24}, \cite{b25} and \cite{b26}.

\subsection{What-If Tool}
The "What If Tool" (WIT) is open-source auditing tool that provides an interactive visualization platform that allows users to probe, visualize, and analyze their data against different classifiers \cite{b27}.  Visualization of hypothetical scenarios can be made against different subsets of inputs and analysis can be assessed related to the importance of different data features over different models.

WIT converts datasets to a numerical format with a graphical display. Each data point is displayed in the graph are editable and can be set to different values to observe counterfactual points in the data. 
What if tool can be useful in identify bias using hypothetical situations against different features using linear classifier and neural network classifier. Models of classification can be assessed against any input features. A ground truth feature can also assess binary categories.

WIT also has a visualization tool called Facets Overview \cite{b28} that uses subsets of the data to compare fairness metrics and assess overall model performance.

WIT with SHAP (SHapley Additive exPlanations)\cite{b29}, as a technique to audit a black-box machine learning model used in the healthcare domain \cite{b30}.

Code and instructions can be found at \cite{b31} and \cite{b32}.

\subsection{FairSearch}
FairSearch is the fair open-source search API to apply fairness in ranked search results \cite{b33}.
Its ranked group fairness definition extends group fairness using the standard notion of protected groups, based on ensuring that the proportion of protected candidates in every prefix of the top-k ranking remains statistically above or indistinguishable from a given minimum \cite{b34}.

Code and instructions can be found at \cite{b35} and \cite{b36}.

\subsection{FairTest}
The FairTest tool helps to identify unwanted associations, unfair, discriminatory, or offensive user treatment in data-driven applications. It examines and biases between potential outcomes and sensitive attributes \cite{b37}. It has been used to address disparate impact, offensive labeling, and uneven
rates of algorithmic error in a variety of datasets.

Code and instructions can be found at \cite{b38}, \cite{b39} and \cite{b40}.

\subsection{Aequitas}
Aequitas is an open-source Python library with many fairness metrics to detect bias \cite{b41}.
To use Aequitas one must upload their data to the tool. Some may not be comfortable in using their data on other sites. If there are data leakage concerns, you should use another tool. 
Aequitas uses a Python library and a command line interface. Users upload their data from the system being audited, configure bias metrics for protected attribute groups of interest as well as reference groups, and then the tool generates bias reports. Bias assessments can be made prior to model selection, evaluating the disparities of the various models trained based on whatever training data was used to tune it for its task. The audits can be performed prior to a model being operationalized or it can be used to determine bias in an A/B testing environment using limited trials and revising the algorithm accordingly.

Code and instructions can be found at \cite{b42}, \cite{b43} and \cite{b44}.

\subsection{Themis-ML}
Themis-ML is a tool that makes model agnostic predictions using reject object recognition (ROC) and discrimination aware ensemble classification (DAEC). The concept is that it that looks for bias by altering the training data by reweighting and relabeling the classes using sampling.
Themis-ML will modify your dataset, so be sure to save your original model and make one or more copies to run on Themis-ML. \cite{b45}

Code and instructions can be found at \cite{b46} and \cite{b47}.

\subsection{Fairness Test-Bed}
The "Fairness-Aware" Machine Learning interface, Fairness Test-Bed, is an open-source framework that facilitates a direct comparison of multiple algorithms \cite{b48} and \cite{b49}.
The framework for the tool focuses on factors like pre-processing if multiple algorithms are going to be compared to each other. The Fairness Test-Bed paper shows that that a combination of class-sensitive error rates and either DI or CV is a good minimal working set. To account for training instability, it also shows that the performance of an algorithm in a single training-test split appears to be insufficient. Accordingly, they recommend reporting algorithm success and stability based on a moderate number of randomized training-test splits.
They use the standard accuracy measures of uniform accuracy, the true positive rate (positive predictive value) and the true negative rate (negative predictive value). They also consider the balanced classification rate, which is a version of accuracy that is unweighted per class.

Code and instructions can be found at \cite{b50} and \cite{b51}.

\subsection{ML-Fairness Gym}
ML-fairness gym is dynamic tool implemented in the environment API from OpenAI Gym that builds simple simulations that explores potential long-run impacts of deploying machine learning-based decision systems in social environments. \cite{b52} and \cite{b53}.
This tool is more unique than some of the other tools because the decisions made by an algorithm at step t affect the next decision it will be asked to make at t+1. The simulations allow exploration of censoring in the observation mechanism, errors from the learning algorithm, and interactions between decision policy and environment dynamics. Assessments can be made on fairness characteristics of certain decision policies based on observed data alone. This can assist with future data collection practices.
Simulations can also be used along with reinforcement learning algorithms.
Although ML-fairness-gym is not an officially supported Google product, its environments implement the environment API from OpenAI Gym.

Code and instructions can be found at \cite{b54} and \cite{b55}.

\subsection{RETAIN}
RETAIN uses a Reverse Time Attention Mechanism to build a clinically interpretable ML model using the EHR (Electronic Health Record) data \cite{b56}. The algorithm uses neural attention so that, given patient records, it can make predictions while explaining how each medical code (diagnosis codes, medication codes, or procedure codes) at each visit contributes to the prediction. The algorithm also treats sequence classification as a special case, which it makes at the last timestep only.
There is a potential trade-off in that the algorithm learns to pay more attention to new information than old information. This should be analyzed on a case-by-case basis.

Code and instructions can be found at \cite{b57}.

\subsection{Knowledge Graphs with Gephi}
A knowledge graph comprises a graph data model ontology of attributes and other data that may provide information about relatedness. Knowledge graphs model topical coherence of information based on an assumption that information from the same context tends to belong together \cite{b58}. Data are often contextual and knowledge graphs utilize a data interchange standard comprising tuples where each tuple can be considered a small graph \cite{b58}. It may also comprise any other associated attributes linked to the data. The primary purpose is classification, clustering, and relatedness that is able to be presented in visual form. The knowledge graph maps heterogeneous types of knowledge. The combined approaches should result in a decrease in disambiguation errors and result in a more discerning clustering of content and inference from the highly heterogeneous data \cite{b59}.
Gephi is a visualization and exploration software platform for all kinds of graphs and networks. It is open-source and free \cite{b60}.

Code and instructions can be found at \cite{b61}.

\subsection{Adversarial Learning with Networkx}
Data may also be run on a CNN and LSTM neural network framework as a generative adversarial network (GAN) model in order to detect anomalous patterns, such as those as previously described by Zenati et al \cite{b62}.  The results can be visualized using T-SNE as one means of assessment.

Data preparation in graph neural network allows for the representation of complex data in non-Euclidean space for complex relationships and inter-dependencies.  Deep learning has been extended to GNNs extract latent representations from images. Data can be converted and manipulated as .ndjson data files, followed by applying a series of 1-dimensional convolutions, followed by LSTM layers, with the sum of the outputs of all LSTM steps fed into an argmax or softmax layer to make a classification decision using known classes as a starting point.

Networkx \cite{b63} is a useful tool with capabilities in graphing, ability to encode attributes to nodes and edges, loading and saving graphs, and ability to convert to other graph formats such DGL \cite{b64}. 
The remaining steps can be run as a progression of models to evaluate the impact of applying additional modeling methods on classification accuracy and similarity.

Code and instructions for GAN-based anomaly detection can be found at \cite{b65}, \cite{b66} and \cite{b67}.

\subsection{IMV-LSTM}
IMV-LSTM uses a mixture attention mechanism in the hidden layers of recurrent neural networks (RNNs) to capture the input variable importance in a multi-variate time series set up \cite{b68}.
Interpretable insights into the data can be found by utilizing RNNs trained on time series with target and exogenous variables, in addition to accurate prediction. The goal of this process is to capture different dynamics in multi-variable time series and distinguish the contribution of variables to the prediction \cite{b69}.

Code and instructions can be found at \cite{b70}.

\subsection{Data Triangulation}
Data triangulation is an older standard statistical analysis used more frequently in the social sciences, but it also has applicability in machine learning models, as it is used to compare multiplicative groupings of data. Instead of utilizing binary pairings, binary pairings may be made, and then compared to other paired metrics. For example, age (as integers or categorically-transformed by decades) can be compared with binary sex or categorically-transformed race enumeration. All of these factors can then be compared to other features in the dataset and cross-correlated and compared. Data triangulation also works well on qualitative data because of the ability to categorically-transform the data and cross-compare the same categorical data over different features or cross-compare it over different categories.

Machine learning and deep learning models of artificial intelligence can incorporate data triangulation in neural networks in the form of mini-batches of additional data to compare signal to noise ratios, even in noisy datasets \cite{b71}. The theory being that the signals will show up consistently across the multiplier iterations of datasets, even if some of the data are incomplete, missing, or excessively noisy. Data triangulation compares qualitative and quantitative data and can be used across all data types and from entirely different sources \cite{b71}. Triangulation is a correlation process that attempts to define relational metrics between data while accounting for noise, overlaps, outliers, and density \cite{b71}. Neural networks can also be triangulated concurrently with other neural networks in order to refine classification, reduce output error, and provide a better explanation of output results (explainable AI) \cite{b72} and \cite{b73}.

\section{Mitigation}
What happens after bias is found? There are choices to be made that can impact data cleaning (inclusion, exclusion, averaging of missing data or parts thereof), data weighting (reweighting), feature engineering (shift, exclude, reweight), and optimization that may disregard tails (include or exclude).
No model is an accurate representation of the real-world. It is merely a representation of an approximation. As such, its degree of accuracy is variable because of inputs and necessary manipulation of data in order to compare apples-to-apples

Balanced or representative data sets are commendable, but full-disclosure if whether the representation is synthetic or real-world is necessary. Balanced datasets are not enough.
Resampling the data can provide a way to make assessments on representation bias and mitigate against it by retraining from the resampled dataset.

Data are what they are. The real-world is messy. Real-world data are messy.
Providing commentary and analysis about bias noted in the underlying data or in the models is very important, especially where the bias can't be mitigated without over-correcting.

\subsection{Caution Against Over-Correcting}
Just as over-fitting and under-fitting are not optimal for analysis, explicit removal of biases and variations are not necessarily appropriate. This type of action is tantamount to introducing new subjective bias (over-correcting) into the data or model, directly impacting outcome. It could lead to a deliberately false model, which would be inherently unethical.
Similarly, caution is warranted when considering introducing additional variables to offset any biasing variables.
Instead, it is recommended to provide an explanation detailing the FATE assessment so that others may understand the limitations of the data or the model.
It is better to convey the understanding that the field uses representative approximations and does not provide absolutes. There is much to be gained through a better understanding of data and models that may show bias that the data, alone, does not indicate.
Similarly, all machine learning models are based on linear algebra and differential equations, as well as numerical rather than categorical values. An assessment of floats is not the same as a contextual assessment of categorical data. However, it is the best approximation that may be available at any given time.

\section{Conclusion and Future Work}
Many people have asked where to find papers and code for auditing tools. This review is a response to those requests.
The goal of this work is to share the information as broadly and as quickly as possible, in light of the ever-changing, challenging environment and the desire of many to want to take proactive measures in their own work for the betterment of the field.

I have endeavored to provide multiple sources of information for each tool including Github links, links to video tutorials, and even PowerPoint slide presentations when the authors made them publicly available.

The tools addressed herein have been utilized on many different datasets by multiple groups all over the world. I have personally used several of them and have found to be reliable in finding bias among real-world and publicly available datasets across multiple domains \cite{b74}. Many of these tools are also model-agnostic.

Many talented researchers and programmers have taken their time to create open-source tools for the greater good. Lets own our ethical algorithmic responsibilities to each other.
Use the tools and pay it forward.

My personal thanks to my AI ethics colleagues and to all of the hardworking researchers who continually strive to promote fundamental fairness across algorithmic domains.

Future work consists of continuing to seek and promote algorithmic fairness, audits, privacy, and human rights in the field of machine learning.

\vspace{12pt}
\end{document}